\documentclass{article}
\usepackage{spconf,amsmath,amssymb,graphicx}
\usepackage[T1]{fontenc}
\usepackage{enumitem}
\usepackage{tabularx}
\setlist{nosep,leftmargin=14pt}

\title{RegionFM: Interpretable Region-Based Brain MRI Classification Using Foundation Model Embeddings}

\name{Wei Zhang, Ming Tang}
\address{L3S Research Center, Hannover, Germany\\
\texttt{wei.zhang@l3s.de, tang@l3s.de}}

\begin{document}
\maketitle

\begin{abstract}
Foundation models provide powerful representations for brain MRI analysis, but their predictions remain difficult to interpret in anatomically meaningful terms. Clinical assessment commonly focuses on anatomical structures and regional abnormalities, whereas conventional explanation methods often produce voxel- or patch-level importance maps that do not directly quantify the contributions of individual brain regions. 
To address this mismatch, we propose RegionFM, an interpretable framework that integrates anatomical segmentation with brain MRI foundation-model embeddings. RegionFM constructs separate MRI volumes for individual anatomical regions or clinically motivated groups and encodes each volume using a frozen foundation model.  
An additive classifier combines these embeddings through explicit regional contributions to the prediction logit, supporting cohort-level explanations.
We evaluate RegionFM using three pretrained encoders on cognitive-impairment classification tasks from OASIS-1 and NACC SCAN. 
The results demonstrate that RegionFM can approach the predictive performance of whole-MRI fine-tuning while providing explicit anatomical explanations.
\end{abstract}


\section{Introduction}

Foundation models \cite{ref_brainiac,ref_neurojepa} have recently emerged as a promising paradigm for medical image analysis. Through pretraining on large-scale imaging datasets, these models learn transferable representations that support downstream tasks such as disease classification and patient stratification. In brain MRI analysis, they can capture complex anatomical and imaging patterns that may be difficult to define manually. However, classifiers built on these representations often operate as black-box systems and provide limited insight into the anatomical evidence underlying their predictions.

Interpretability is especially important for brain MRI because clinical assessment is commonly organized around anatomically defined structures and regional abnormalities. For example, clinicians may examine regional atrophy, cortical thickness, ventricular enlargement, lesion burden, or white-matter hyperintensities when evaluating neurological conditions. In contrast, conventional explanation methods often produce voxel- or patch-level importance maps. Although these maps can localize predictive evidence, they do not directly quantify how individual anatomical structures contribute to a prediction. Consequently, the unit of explanation does not align with the anatomy-based reasoning used in clinical assessment. This motivates a framework in which anatomical regions serve directly as units of both prediction and explanation.

To address this need, we propose RegionFM, an interpretable region-based framework for brain MRI classification using foundation-model embeddings. RegionFM first applies anatomical segmentation to construct separate MRI volumes for individual brain regions or clinically motivated groups. A frozen foundation model encodes each volume into an embedding, and an additive classifier maps each embedding to an explicit scalar contribution to the prediction logit. The final logit is the sum of the available anatomical contributions and a global bias. These contributions can be aggregated across participants to identify anatomical patterns associated with the model’s predictions, supporting cohort-level explanations.

We evaluate RegionFM using BrainIAC, Neuro-JEPA, and Neuro-JEPA-Large on two cognitive-impairment classification tasks from OASIS-1~\cite{ref_oasis_1} and NACC SCAN~\cite{ref_nacc}. We compare individual atlas regions with coarser anatomical groups and assess predictive performance against whole-MRI linear probing and fine-tuning. 
Beyond classification performance, we analyze cohort-level anatomical contribution differences between positive-class and control participants using the best-performing RegionFM configuration, and compare the resulting patterns with previous neuroimaging findings.

Our main contributions are as follows:
\begin{itemize}
    \item We propose RegionFM, an anatomically structured framework that combines frozen foundation-model embeddings with an additive classifier, providing explicit regional contributions to predictions and supporting cohort-level explanations.

    \item We systematically compare individual regions and clinically motivated anatomical groups across three pretrained encoders and two datasets, examining how anatomical scale affects predictive performance.

    \item We analyze cohort-level anatomical contributions of the best-performing RegionFM configuration and compare the resulting patterns with prior neuroimaging findings.
\end{itemize}

\begin{figure*}[t]
    \centering
    \includegraphics[width=\textwidth]{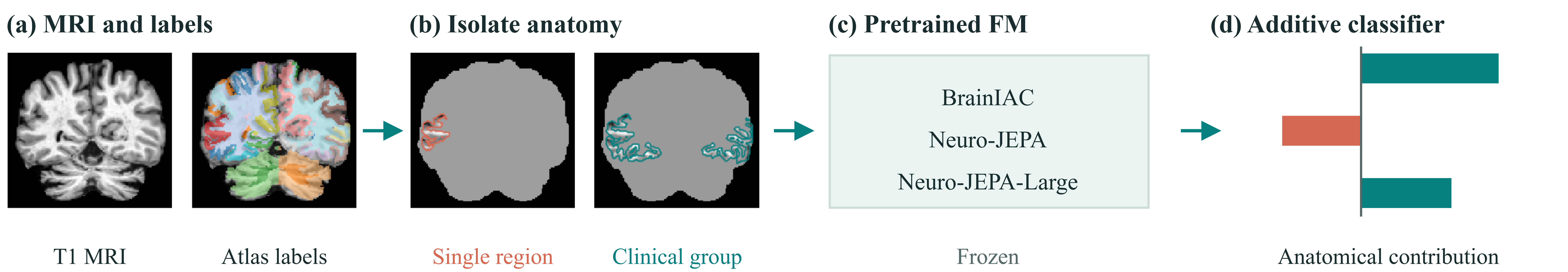}
    \caption{Overview of RegionFM. Each MRI is anatomically segmented to construct region- or group-specific volumes, which are encoded by a frozen foundation model. An additive classifier combines the embeddings and decomposes the prediction into explicit anatomical contributions.}
    \label{fig:regionfm_framework}
\end{figure*}

\section{Related Work}

Brain MRI foundation models learn transferable representations through large-scale pretraining. BrainIAC uses contrastive learning and evaluates adaptation across neurological tasks, including settings with limited labeled data~\cite{ref_brainiac}. Neuro-JEPA combines a latent predictive objective with a sparse mixture-of-experts architecture for structural MRI sequences~\cite{ref_neurojepa}. These models support downstream classification through frozen whole-MRI representations or encoder fine-tuning. RegionFM uses these pretrained encoders on anatomically isolated volumes to obtain separate representations for individual regions or groups.

Anatomical MRI classification has long used regional measurements and spatially organized features. Cuingnet et al.~\cite{ref_cuingnet} compared voxel-based, cortical-thickness, and hippocampal approaches for Alzheimer’s disease (AD) and Mild Cognitive Impairment (MCI) classification. Liu et al.~\cite{ref_ldmil} extended local feature learning through landmark-based deep multi-instance learning, combining anatomical image patches and their joint configuration. RegionFM uses atlas-defined units and explicitly compares individual regions with coarser anatomical groups, examining how anatomical scale affects prediction and explanation.

Additive models provide explicit contributions through sums of input-specific functions~\cite{ref_gam,ref_nam}. Anatomical explanations have also been obtained by aggregating post-hoc voxel relevance within brain regions~\cite{ref_bohle}. RegionFM applies additive modeling to anatomical foundation-model embeddings, making each regional contribution an explicit term used to compute the prediction logit. Its contribution lies in this combination of pretrained representations, anatomical decomposition, and additive prediction.

\section{Method}

Given an MRI scan, RegionFM obtains anatomical labels, constructs one volume for each anatomical unit, extracts a foundation-model embedding per unit, and performs classification with an additive model (Fig.~\ref{fig:regionfm_framework}).

\subsection{Anatomical Units}

An anatomical parcellation is obtained from FreeSurfer or FastSurfer~\cite{ref_freesurfer,ref_fastsurfer}, defining anatomical units $\mathcal{R}=\{R_1,\ldots,R_K\}$, where $K$ depends on the dataset and strategy. RegionFM-Regions retains individual atlas labels, while RegionFM-Clinical-15 combines related labels into anatomical groups before encoding.

The 15 anatomical groups are: left medial temporal; right medial temporal; posterior cingulate and precuneus; lateral temporal; parietal (excluding precuneus); frontal; anterior cingulate; insula; occipital; deep gray matter; ventricles and cerebrospinal fluid; cerebral white matter; white-matter hypointensities; cerebellum; and brainstem.

NACC SCAN contains 95 individual labels, all assigned to the 15 groups. OASIS-1 contains 110 individual labels, of which 102 enter clinical grouping, forming 14 groups because the insula group is not represented in its official segmentation labels. These counts describe cohort-level inventories and individual participants may have unavailable units. Label maps are transformed into the foundation-model input space.

\subsection{Region-Wise MRI Construction}

Let $X_i$ be the preprocessed MRI, $B_i$ the subject's binary brain mask, and $M_{i,k}$ the binary mask of anatomical unit $R_k$ restricted to $B_i$. For each nonempty target mask, we calculate $c_{i,k}$ as the mean intensity of its voxels and construct
\begin{equation}
\begin{aligned}
X_{i,k} ={}& M_{i,k}\odot X_i
          +(B_i-M_{i,k})c_{i,k}\\
          &+(1-B_i)\odot X_i.
\end{aligned}
\label{eq:masked_mri}
\end{equation}
Target voxels and non-brain background remain unchanged, while other brain voxels are replaced with the target unit's mean intensity. For a clinical group, the constituent region masks are combined before calculating the mean and constructing the volume. Thus, $c_{i,k}$ depends on both the subject and the anatomical unit.

\subsection{Frozen Foundation-Model Encoding}

Each unit-specific volume is encoded by a frozen pretrained model $f_{\theta}$:
\begin{equation}
\mathbf{z}_{i,k}=f_{\theta}(X_{i,k}), \qquad \mathbf{z}_{i,k}\in\mathbb{R}^{d}.
\end{equation}
The subject is represented by $Z_i=\{\mathbf{z}_{i,1},\ldots,\mathbf{z}_{i,K}\}$. Freezing $f_{\theta}$ keeps optimization focused on the additive prediction layer and permits the same classifier design across encoders.

\subsection{Mask-Aware Region-Additive Classification}

For each unit $k$ and embedding feature $j$, we estimate a mean $\mu_{k,j}$ and population standard deviation $\sigma_{k,j}$ using only training subjects for whom that unit is available. For available units, we standardize each feature as
\begin{equation}
\tilde z_{i,k,j}=\frac{z_{i,k,j}-\mu_{k,j}}{\max(\sigma_{k,j},10^{-6})}.
\label{eq:standardization}
\end{equation}
These statistics are fitted separately for each seed and applied unchanged to validation and test subjects. Unavailable embeddings are excluded when fitting the statistics and set to zero after standardization.

Each anatomical unit has a bias-free linear head shared across subjects:
\begin{equation}
s_{i,k}=\mathbf{w}_k^{\top}\tilde{\mathbf{z}}_{i,k}.
\label{eq:region_contribution}
\end{equation}

Let $a_{i,k}\in\{0,1\}$ indicate whether unit $k$ is available according to the criteria in Experimental Setup. The prediction logit and positive-class probability are
\begin{equation}
h_i=b_0+\sum_{k=1}^{K}a_{i,k}s_{i,k},
\qquad
\hat y_i=\sigma(h_i),
\label{eq:subject_logit}
\end{equation}
where $\sigma(t)=(1+e^{-t})^{-1}$ is the sigmoid function. The regional weights $\mathbf{w}_k$ and global bias $b_0$ are trainable, while the foundation-model encoder remains frozen. No regional biases are used. Availability masking forces unavailable units to contribute exactly zero to the prediction. The model is trained with binary cross-entropy.

\subsection{Interpretation}

Equation~\ref{eq:subject_logit} provides an exact decomposition of the prediction logit into anatomical contributions and a global bias. For available units, positive contributions increase the logit toward the positive class, whereas negative contributions decrease it toward the control class. Aggregating masked contributions $a_{i,k}s_{i,k}$ across participants supports cohort-level anatomical explanations. These contributions describe evidence used by the model.

\begin{table}[t]
\centering
\caption{Predictive performance measured by test ROC-AUC. Bold denotes the best whole-MRI and RegionFM result in each dataset.}
\label{tab:main_results}
\small

\resizebox{\columnwidth}{!}{

\begin{tabular}{llcc}
\hline
\textbf{Encoder} & \textbf{Method} & \textbf{OASIS-1} & \textbf{NACC SCAN} \\
\hline
BrainIAC & Whole-MRI linear probe & $0.746\pm0.081$ & $0.721\pm0.024$ \\
BrainIAC & Whole-MRI FT & $0.736\pm0.086$ & $0.793\pm0.037$ \\
BrainIAC & RegionFM-Regions & $0.626\pm0.124$ & $0.675\pm0.023$ \\
BrainIAC & RegionFM-Clinical-15 & $0.616\pm0.090$ & $0.746\pm0.023$ \\
\hline
Neuro-JEPA & Whole-MRI linear probe & $0.683\pm0.101$ & $0.925\pm0.023$ \\
Neuro-JEPA & Whole-MRI FT & $\mathbf{0.760\pm0.086}$ & $\mathbf{0.938\pm0.023}$ \\
Neuro-JEPA & RegionFM-Regions & $0.556\pm0.173$ & $0.659\pm0.025$ \\
Neuro-JEPA & RegionFM-Clinical-15 & $\mathbf{0.730\pm0.112}$ & $\mathbf{0.848\pm0.017}$ \\
\hline
Neuro-JEPA-Large & Whole-MRI linear probe & $0.665\pm0.113$ & $0.917\pm0.023$ \\
Neuro-JEPA-Large & Whole-MRI FT & $0.734\pm0.075$ & $0.894\pm0.013$ \\
Neuro-JEPA-Large & RegionFM-Regions & $0.545\pm0.109$ & $0.672\pm0.033$ \\
Neuro-JEPA-Large & RegionFM-Clinical-15 & $0.705\pm0.075$ & $0.844\pm0.007$ \\
\hline
\end{tabular}

}
\end{table}

\section{Experimental Setup}

\subsection{Datasets and Tasks}

We evaluate RegionFM on two brain MRI classification tasks. OASIS-1~\cite{ref_oasis_1} includes 226 participants with available official FreeSurfer data: 133 controls with Clinical Dementia Rating (CDR) $=0$ and 93 participants with cognitive impairment defined by CDR $>0$.
NACC SCAN~\cite{ref_nacc} includes one T1-weighted scan per participant, with 552 controls and 552 participants with AD dementia.
Both datasets are partitioned into training, validation, and test sets using five random seeds. All methods use identical partitions within each dataset and seed.

\subsection{Preprocessing and Models}

All encoders receive the same BrainIAC-preprocessed MRI within each cohort. OASIS-1 uses the official FreeSurfer parcellations, whereas NACC SCAN uses FastSurfer~\cite{ref_freesurfer,ref_fastsurfer}. Anatomical labels are mapped into the preprocessed MRI space using nearest-neighbor interpolation.

An anatomical unit is available if it overlaps the brain mask by at least one voxel and its embedding is successfully extracted. The brain mask comprises finite, nonzero voxels of the preprocessed MRI. For Clinical-15, at least 50\% of the constituent labels in the dataset-specific group map must overlap this mask. Unavailable units contribute zero to the prediction. Units without available training embeddings are disabled in all splits for that seed.

We evaluate BrainIAC~\cite{ref_brainiac}, Neuro-JEPA, and Neuro-JEPA-Large~\cite{ref_neurojepa}. For each encoder, RegionFM uses frozen embeddings extracted from individual atlas regions or Clinical-15 groups. Whole-MRI baselines include linear probing on frozen embeddings and fine-tuning of the encoder and classification head. Both Neuro-JEPA variants use the no-bias-update fine-tuning configuration.

\subsection{Training and Evaluation}

RegionFM selects the checkpoint with the lowest validation loss and uses early stopping with a patience of 20 epochs. Linear probes select checkpoints by validation ROC-AUC, while whole-MRI fine-tuning follows the corresponding encoder's training protocol. Test sets are reserved for final evaluation. Predictive performance is measured by test ROC-AUC, reported as mean $\pm$ standard deviation across the five seeds.

\section{Results}

\subsection{Predictive Performance}

Table~\ref{tab:main_results} compares RegionFM with whole-MRI linear probing and fine-tuning. Among the RegionFM configurations, Clinical-15 with Neuro-JEPA achieves the highest mean test ROC-AUC on both datasets: 0.730 on OASIS-1 and 0.848 on NACC SCAN. The corresponding whole-MRI fine-tuning results are 0.760 and 0.938, respectively. Thus, grouped RegionFM approaches the predictive performance of fine-tuning on OASIS-1 and NACC SCAN while providing explicit anatomical contributions. Consistent with its reported advantage over BrainIAC~\cite{ref_neurojepa}, Neuro-JEPA achieves higher mean ROC-AUC on both datasets with whole-MRI fine-tuning and RegionFM-Clinical-15.

Anatomical grouping substantially improves performance for both Neuro-JEPA encoders. With Neuro-JEPA, replacing individual regions with Clinical-15 groups increases ROC-AUC from 0.556 to 0.730 on OASIS-1 and from 0.659 to 0.848 on NACC SCAN. Neuro-JEPA-Large follows the same pattern. For BrainIAC, however, grouping improves performance on NACC SCAN but slightly decreases it on OASIS-1. The benefit of grouping therefore varies across encoders and datasets.

Compared with frozen whole-MRI linear probing, Clinical-15 achieves higher mean ROC-AUC on OASIS-1 for both Neuro-JEPA variants, but lower ROC-AUC on NACC SCAN. Whole-MRI representations therefore remain a strong baseline even without encoder fine-tuning. Increasing encoder size also does not improve mean ROC-AUC in the grouped setting: Neuro-JEPA-Large achieves 0.705 on OASIS-1 and 0.844 on NACC SCAN, compared with 0.730 and 0.848 for Neuro-JEPA.

\subsection{Cohort-Level Anatomical Patterns}

\begin{figure}[!t]
\centering
\includegraphics[width=\columnwidth]{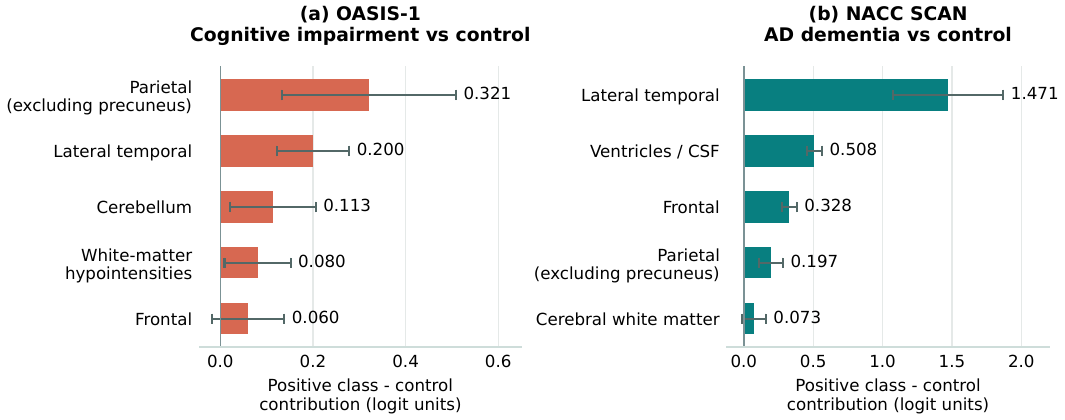}
\caption{Top-five anatomical contribution differences for Neuro-JEPA Clinical-15 RegionFM (positive class minus control). The numbers show mean $\pm$ standard deviation across five seeds. Values are in logit units, horizontal scales differ.}
\label{fig:neurojepa_cohort_patterns}
\end{figure}

We focus here on RegionFM-Clinical-15 with the Neuro-JEPA encoder, which achieves the highest mean test ROC-AUC among the evaluated RegionFM configurations on both datasets. Within each seed, we compute each group's mean masked contribution among positive-class test participants minus its mean contribution among controls, then average these differences across seeds. All test participants are included according to their true class labels, irrespective of prediction correctness. Figure~\ref{fig:neurojepa_cohort_patterns} shows the five groups with the largest positive mean differences in each cohort.

On OASIS-1, parietal (excluding precuneus), lateral temporal, and cerebellar groups have the largest positive mean differences, followed by white-matter hypointensities and frontal regions. On NACC SCAN, lateral temporal and ventricular/CSF groups lead, followed by frontal, parietal (excluding precuneus), and cerebral white matter. Lateral temporal, parietal, and frontal groups therefore appear among the top five in both cohorts, although their ordering differs.

These shared cortical findings are anatomically consistent with previous MRI studies. McDonald et al.~\cite{ref_mcdonald2009} reported elevated atrophy rates in middle and inferior lateral temporal and inferior parietal regions in MCI, with broader parietal and frontal involvement at greater levels of impairment. The ventricular/CSF finding in NACC also has relevant precedent: ventricular expansion has been associated with cortical atrophy in MCI and AD~\cite{ref_madsen2015}. These correspondences support the anatomical plausibility of the highlighted groups, without directly validating the exact Clinical-15 grouping.

The cerebellar and white-matter findings require more qualified interpretation. Cerebellar volume has been associated with baseline cognition in MCI, but not subsequent cognitive change~\cite{ref_lin2020}. Another study found accelerated cerebellar atrophy in AD and MCI progressing to AD, but not stable MCI~\cite{ref_tabatabaei2017}. T1-visible white-matter abnormalities have also been linked to ventricular expansion and potentially age- and vascular-related damage~\cite{ref_coutu2016}, making them nonspecific to AD. These findings provide context for the white-matter contributions but do not establish the biological process represented by either the hypointensity group or the broader cerebral white-matter group.

\section{Conclusion}

We introduced RegionFM, an interpretable framework that combines frozen brain MRI foundation-model embeddings with an additive classifier to provide explicit anatomical contributions and cohort-level explanations. Evaluations on OASIS-1 and NACC SCAN show that clinically grouped Neuro-JEPA provides the strongest RegionFM performance and can approach whole-MRI fine-tuning.
And the literature supports the findings of the resulting anatomical patterns.
RegionFM offers a practical approach to anatomically interpretable MRI classification while making the associated predictive trade-off explicit.



\bibliographystyle{IEEEbib}
\bibliography{refs}

\end{document}